\pdfoutput=1 
\documentclass[11pt,a4paper]{article}
\usepackage[hyperref]{eacl2021}
\usepackage{times}
\usepackage{latexsym}

\usepackage{microtype}
\usepackage{amsmath}
\usepackage{algorithm}
\usepackage[noend]{algpseudocode}
\usepackage{epsfig,graphicx,verbatim,parskip,tabularx,setspace,xspace}
\usepackage{xcolor}
\usepackage{tikz}
\usepackage{subcaption}
\usetikzlibrary{positioning,arrows.meta}

\aclfinalcopy 

\title{Incremental Beam Manipulation for Natural Language Generation}

\author{James Hargreaves \\
  The Trade Desk \& University of Cambridge \\
  \texttt{james.hargreaves@thetradedesk.com} \\\AND
  Andreas Vlachos \\
  University of Cambridge \\
  \texttt{av308@cam.ac.uk} \\ \And
  Guy Emerson \\
  University of Cambridge \\
  \texttt{gete2@cam.ac.uk} \\}

\date{}

\begin{document}
\maketitle
\begin{abstract}
 The performance of natural language generation systems has
 improved substantially with modern neural networks.
 At test time they
 typically employ beam search to avoid locally optimal but globally suboptimal predictions.
 However, due to model errors, a larger beam size can lead to deteriorating performance
 according to the evaluation metric. 
 For this reason, it is common 
 to rerank the output of beam search, but this relies on beam search to produce a good set of hypotheses,
 which limits the potential gains.
 Other alternatives to beam search require changes to the training of the model, which restricts their applicability compared to beam search. This paper proposes incremental beam manipulation, i.e.\
 reranking the hypotheses in the beam during decoding instead of only at the end. This way, hypotheses that are unlikely to lead to a good final output are discarded, and in their place hypotheses that would have been ignored will be considered instead. Applying incremental beam manipulation leads to an improvement of 1.93 and 5.82  BLEU points over vanilla beam search for the test sets of the E2E and WebNLG challenges respectively. The proposed method also outperformed a strong reranker by 1.04 BLEU points on the E2E challenge, while being on par with it on the WebNLG dataset.

\end{abstract}

\section{Introduction}
\label{sec:intro}
In natural language generation (NLG), the goal is to generate text representing structured information (e.g.\ a database record or a meaning representation) that is both fluent and contains the right information. 
%
%
Sequence-to-sequence models (seq2seq) have been effective on many tasks in NLG (for example: \citealp{wen2015stochastic, dusek-jurcicek-2016-sequence}). 
These systems first create an embedding for the input information. 
This embedding is used incrementally during decoding,
generating one token at a time.
Seq2seq models are generally decoded using beam search, to mitigate the effect of locally optimal but globally suboptimal decisions made by greedy search.


The performance of NLG systems can plateau or even decrease when beam sizes larger than 10 are used, which is counter-intuitive since larger beams produce more likely sequences according to the model.
For example, 
\citet{dusek-jurcicek-2016-sequence} used a beam size of 10, and
\citet{asghar2017deep} found a size of 5 to be optimal. 
%
Decreasing performance has been found across a range of tasks including \citep{cohen2019seq}. Moreover, it
and was given by \citet{koehn2017six} as one of the six main challenges facing neural machine translation.
To investigate this, \citet{Stahlberg2019OnNS} presented an exact search algorithm
to find the most likely output according to a seq2seq model.
However, this performed poorly compared to beam search,
demonstrating that search errors (from beam search) can mask model errors (from the seq2seq model).

To mitigate the limitations of beam search, it is common practice to apply a reranker to the final set of hypotheses.
This can be done by defining a reranking criterion
(for example: \citealp{kumar2004mbr,blain2017exploring,borgeaud2020vote})
or by training a reranker to predict the best hypothesis in a beam
(for example: \citealp{dusek-jurcicek-2016-sequence,agarwal2018char2char}).
Training a reranker allows us to take into account information from outside the model and mitigate model errors.
However, rerankers can only choose a hypothesis from the final beam,
which limits their potential.
\begin{figure}
    \centering
    \includegraphics[width=7.7cm]{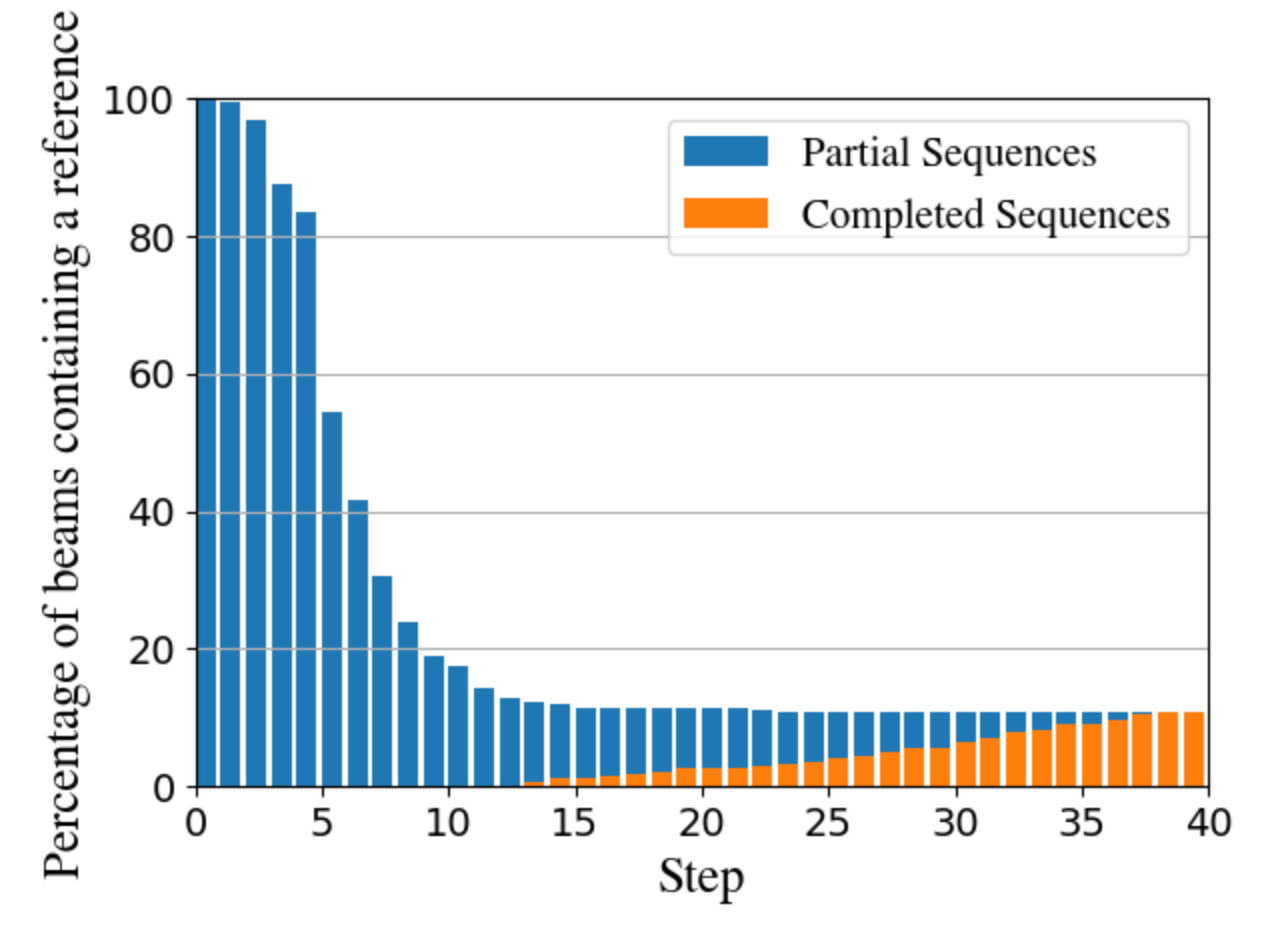}
    \caption{The percentage of beams which
    contain a reference (orange),
    or which could still lead to a reference (blue),
    using the model of \citet{dusek-jurcicek-2016-sequence}
    with beam size 3 on the E2E validation set.
    }
    \label{fig:fallout_location}
\end{figure}

To quantify this, we trained the seq2seq model proposed by \citet{dusek-jurcicek-2016-sequence},
and applied it to the E2E validation set \citep{novikova2017e2e}.
For each instance, we recorded the point at which all gold-standard references fell out of the beam,
meaning that none of the partial hypotheses in the beam could be extended to a gold reference.
A final beam containing at least one of the references would score optimally with an oracle reranker
(providing an upper bound on performance).
Figure~\ref{fig:fallout_location}
shows the results for beam size 3.\footnote{
    For larger beam sizes, the same general trends were observed. See Appendix~\ref{sec:app_fallout} for beam size 10.
}
The final beam contained a reference in only 60 out of 547 cases (11\%). For the remaining 89\% of the cases, even an oracle reranker would be unable to give optimal results. 
The figure also shows that
in over half of the cases,
all references fell out in the first 6 steps.
In contrast, references that were still in the beam at step 15 were almost certain to stay in the beam until the end.
These observations suggest that an \textit{early} manipulation of the beam has a strong potential to improve performance.

In this paper, we propose a method for
manipulating which items are pruned from the beam at each stage of decoding.
We then present evidence that this is a successful approach: it led to an improvement of 1.93, and 5.82 BLEU points over vanilla beam search on the E2E and WebNLG challenges, respectively. When comparing to a strong reranker, the performance of incremental beam manipulation was similar on the WebNLG dataset, whilst increasing the performance on the E2E challenge by 1.04 points.
We also applied beam manipulation on top of 
length normalisation \cite{murray2018correcting}, and incremental beam manipulation was able to improve its performance.


\section{Related Work}
\label{sec:relwork}

This paper is far from the first to try to improve beam search for natural language generation. 

One modification is to use a variable beam size instead of a fixed one \citep{freitag-al-onaizan-2017-beam}.
However, this can only improve decoding speed, as the ranking of the hypotheses in the beam remains unchanged, and thus model errors are exposed by the reduction of search errors.


Length normalisation \cite{murray2018correcting} is widely used strategy that often improves the performance of a beam search decoder, by mitigating the fact that seq2seq models are biased towards generating shorter sequences. Rather than directly using model probabilities to order the hypotheses in the beam, each probability is normalised according to the length of the hypothesis, so that shorter hypotheses are penalised. However, this only has an impact once the hypotheses within the beam have different lengths. This only occurs towards the end of the decoding process, and we showed in the previous section that the reference hypotheses often fall out of the beam relatively early. Furthermore, \citet{Stahlberg2019OnNS} showed that the biases causing the deteriorating model performance are more complex than a simple length bias.

\citet{wiseman2016sequence} modified the training procedure for seq2seq models. They ran beam search and introduced a loss each time the gold standard sequence fell out of the beam. \citet{goyal2018continuous} and \citet{collobert2019diffbeam} also modified the training procedure. They added a term to the loss function that approximated the loss that the model would receive when generating using a beam search method for each example in the training set.
However, one of the reasons that beam search has been so widely used is that it can be applied on top of a language model without changing the training procedure, and this is lost with these approaches.

\citet{gu2017trainable} manipulated the hidden state of the language model at each step of the decoding. This was achieved via a multi-output regressor that produced a vector that is added to the hidden state used in decoding.
The regressor was trained via reinforcement learning, and the training signal was gathered by injecting unstructured noise to the hidden state.
\citet{chen2018stable} also manipulate the hidden state.
For each training instance, they apply beam search and take the hypothesis with the highest BLEU score. The manipulator network is trained to encourage a greedy decoder to produce this output. 
Both of these approaches rely on inferring a better hidden state to be used in the decoding, which is not straightforward to define. We instead manipulate the hypotheses in the beam directly.


Finally, \citet{negrinho2018learning} presented a framework for learning a beam search framework via imitation learning. This resulted in a beam aware algorithm which was proved to have no regret guarantees. While this paper makes a compelling argument for this method in theory,
putting it into practice requires a number of further engineering decisions.
Our work can be seen as a way of applying this general framework using a simple and computationally efficient roll-out strategy.

\section{Incremental Beam Manipulation}


In order to describe our method for incremental beam search, we first introduce terminology to describe a standard beam search decoder. The decoder produces a sequence iteratively, token by token.
At each iteration it performs 3 actions:  expand, rank and prune. The expand step generates all possible next step hypotheses. The rank step orders these hypotheses from most likely to least likely. The pruning step then removes the hypotheses that are near the end of this order.

This formulation of the beam search algorithm enables us to view beam manipulation as a ranking problem since expand is determined by the (fixed) decoder and the size of the beam chosen determines the pruning. The rank step determines which hypotheses will not be kept in the next iteration
and hence discarded. By modifying the ranking method used, we can choose the partial hypotheses expanded during beam search, taking into account the current state of the beam as well as signals beyond model scores.

It is worth noting that while this paper 
applies beam manipulation on top of a seq2seq model, the techniques used could be applied without change to any conditional or unconditional neural language model that can be decoded using beam search.


\subsection{Ranking via Roll-out}

Partial hypotheses are 
more difficult to rank than complete hypotheses since the rest of the generated text is unknown. For example, consider the following partial hypotheses:
\begin{quote}
    Loch Fyne is a restaurant located...
\end{quote}
\begin{quote}
    There is a family friendly...
\end{quote}

Both of these convey some information about a family-friendly restaurant named `Loch Fyne'.
It is hard to know which partial sequence will lead to a better complete sentence,
which is what we would like a ranker to tell us.
Existing rerankers often rely on detecting missing information, but some information may still be to come for partial hypotheses.


What we need is a way to rank partial hypotheses
based on how the seq2seq model is likely to complete them.
We propose ranking partial hypotheses based on a \emph{greedy roll-out}.
This is a computationally efficient approximation of how the seq2seq model might complete the partial hypothesis.

In the existing literature, roll-outs are generally used at training time \cite{krishnamurthy2015learning}, for the situation where the model's subsequent decisions influence the loss function for an individual decision. The roll-outs are used to produce an approximation to the final sequence that would be reached if the original action was taken. This enables a value for the loss of the original decision to be predicted.

On the other hand, incremental beam manipulation aims to predict which partial hypotheses will lead to good completed sequences. Similar to traditional roll-outs, this is impacted by the generating model's subsequent decisions. In this case, the difference is that roll-outs are used to provide features in addition to obtaining training signal. It is also worth noting that for incremental beam manipulation we use roll-outs at test time as well as at training time. 

\begin{figure*}
\centering
\begin{tikzpicture}[very thick, execute at begin node=\strut]
	\tikzstyle{state}=[draw=black, rounded corners, font=\small\tt, inner sep=1mm]
	\tikzstyle{rollout}=[state, dashed]
	\tikzstyle{label}=[font=\bf]
	\tikzstyle{arrow}=[-{Straight Barb[round,angle=60:2mm]}, shorten >=.5mm]
	\tikzstyle{column 1}=[anchor=east]
	\tikzstyle{column 2}=[anchor=west]
	\tikzstyle{column 3}=[label]
	\tikzstyle{row 1}=[label]
	\matrix[row sep=3mm, column sep=0mm] {
		\node {Current beam};
		&[8mm] \node {Next token and greedy roll-out};
		& \node {Rank}; \\
		\node[state] (beam1) {Loch\_Fyne is a};
		& \node[state] (beam1expand1) {family};
			\node[rollout, anchor=west] at (beam1expand1.east)
				{friendly restaurant in the city centre <E>};
		& \node {1}; \\
		& \node[state] (beam1expand2) {restaurant};
			\node[rollout, anchor=west] at (beam1expand2.east)
				{located in the centre of the city <E>};
		& \node {3}; \\
		\node[state] (beam2) {There is a};
		& \node[state] (beam2expand1) {restaurant};
			\node[rollout, anchor=west] at (beam2expand1.east)
				{in the city centre called Loch\_Fyne <E>};
		& \node {4}; \\
		&\node[state] (beam2expand2) {child};
			\node[rollout, anchor=west] at (beam2expand2.east) (beam2rollout2)
				{friendly restaurant called Loch\_Fyne <E>};
		& \node {2}; \\
	};
	\draw[arrow] (beam1.east) -- (beam1expand1.west);
	\draw[arrow] (beam1.east) -- (beam1expand2.west);
	\draw[arrow] (beam2.east) -- (beam2expand1.west);
	\draw[arrow] (beam2.east) -- (beam2expand2.west);
\end{tikzpicture}
\caption{%
	Incremental beam manipulation, for a beam size of 2, when generating the 4th token.
	Each element of the beam is expanded,
	and each partial hypothesis is greedily rolled out to a complete hypothesis.
	The complete hypotheses are ranked and then pruned.
	The partial hypotheses are then used in the next step of beam search.
}
\label{fig:manipulate}
\end{figure*}
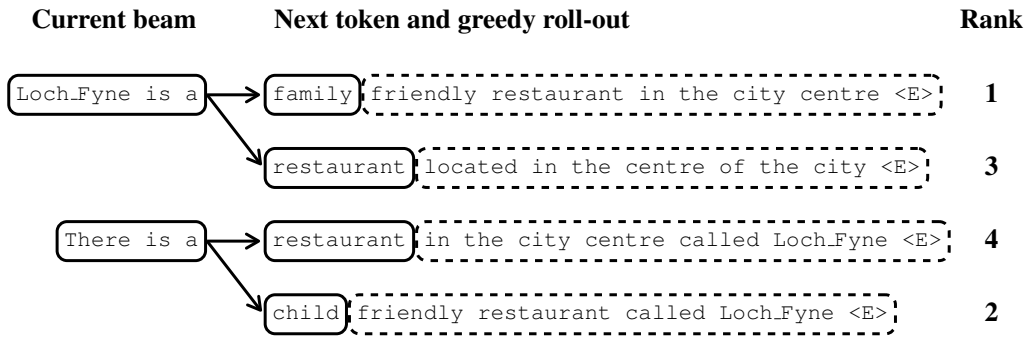

Beam manipulation can be applied after any step in the beam search decoding. Figure~\ref{fig:manipulate} illustrates a single manipulation. The roll-outs are used to produce approximations to the completed hypotheses that would be produced if the partial hypothesis remained in the beam. These completed sequences are then ranked to define an order of the partial sequences. Since this may result in different hypotheses remaining in the beam, the area of the search space considered during the decoding has been manipulated.


\subsection{Reranker Architecture}
\label{sec:IR_rankers}
Incremental beam manipulation requires a method to rank completed hypotheses. There are many existing rerankers designed for this task, such as the TGEN reranker \cite{dusek-jurcicek-2016-sequence}. However, these rerankers are unlikely to be effective when used in incremental beam manipulation.
In the latter, the partial hypotheses need to be ranked according to their potential to produce good completed hypotheses. This is a related but different task to that of a traditional reranker that aims to identify the hypotheses that will score best against some metric such as BLEU score. Rerankers of completed hypotheses typically rely on input information missing from the output as the signal; however, this is not necessarily useful when reranking partial hypotheses.
For example, it is more useful to identify partial hypotheses which are indicative of model failure at an early stage of decoding.

%

To rank the partial hypotheses via roll-outs as introduced in the previous section, we explored two commonly used techniques in the field of information retrieval: pointwise ranking and pairwise ranking \citep{learning-to-rank-for-ir}. Pointwise approaches predict a numerical value for each item, and the items are ordered by sorting them according to this value. Pairwise approaches, given a pair of hypotheses, output which of them would rank more highly,
and techniques such as the Copeland method \citep{copeland_method} are then used to produce a total ordering from these pairwise comparisons.
In preliminary experiments, the pointwise approach outperformed the pairwise approach. These results are summarised in Appendix~\ref{sec:app_point_pair}. For the remainder of this paper, we will focus on the pointwise approach.

The inputs to the reranker were:
\begin{itemize}
    \item The meaning representation (i.e.\ the structured information input for NLG). This was passed as a sequence.
    \item The generated text produced by the roll-out of the partial hypothesis. This was passed as a sequence surrounded by the start token \texttt{<S>} and end token \texttt{<E>}.
    \item The rank, according to the seq2seq model probability, of the completed sequence in the beam. Since this was a categorical variable, it was passed as a one-hot vector.
\end{itemize}

\begin{figure}
\includegraphics[width=7.7cm]{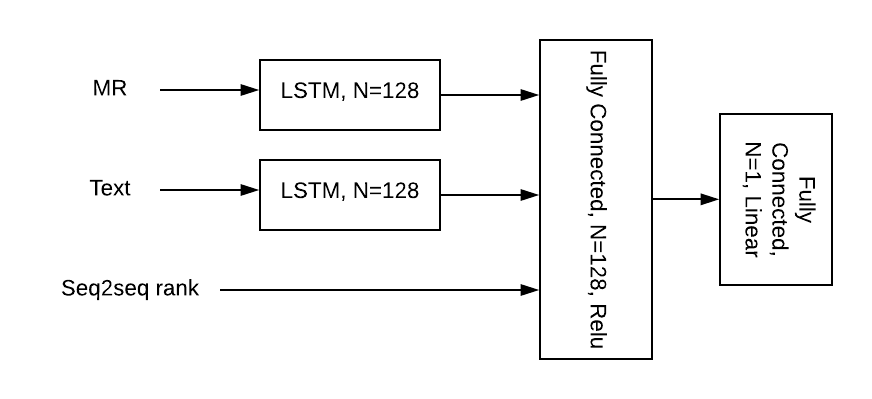}
\caption{\label{fig:architectures}Architecture of the reranker. N~is the number of hidden units for the LSTMs and the number of output nodes for the fully connected layers. For the fully connected layers, the activation function is shown.}
\end{figure}

The architecture of the reranker is summarised in Figure~\ref{fig:architectures}. This model is used to assign a value to each hypothesis individually. The meaning representation and the rolled-out text are each passed through an LSTM,\footnote{%
    For efficient batching, it is common practice to add padding tokens to make sequences the same length. We prepended padding tokens rather than appending them, as this led to better performance in preliminary experiments. 
    This is presumably because prepended tokens have a smaller impact on the final hidden state.
}
and then all inputs are concatenated and passed through two fully connected layers.

At inference time the reranker is used to identify poorly performing hypotheses so that these can be pruned. This enabled the task of the reranker to be simplified to distinguishing between those hypotheses near the bottom of the beam from the hypotheses in the rest of the beam. Therefore, we used the reranker's scores to split the hypotheses into two groups: those with scores in the bottom quartile and those with scores in the top three quartiles. The hypotheses within each group were ordered by the seq2seq model's probability.

In preliminary experiments,
we also tried using the reranker to provide a total ordering.
However, using it to provide a coarse partial ordering (as described above)
gave more promising results.





\subsection{Training the Reranker}
\label{sec:training_ranker}

We trained the reranker on completed hypotheses that were ranked from best to worst.
The sequences were produced by generating text sequences using the seq2seq model. A beam search decoding, with a large beam size, was applied to each instance in the training set. \footnote{ In preliminary experiments, we held back a portion of the training set from the seq2seq model's training phase, so that the beam manipulation ranker was trained on outputs that the seq2seq model had not seen. However, the system was unable to recover from the lower performance of the seq2seq model.}
The set of hypotheses present in the final beam of the search was ranked from best to worst and recorded.

The notion of the best hypothesis was simplified to the one that received the highest BLEU score \cite{papineni-etal-2002-bleu} against the manually written references. 
BLEU was chosen due to its wide adoption as an automatic evaluation measure, but any automatic metric could have been used in its place. 

As discussed at the end of the previous section, we need the reranker to distinguish between hypotheses that should be pruned from those that should be kept.
Furthermore, for the purpose of reranking, only relative differences in BLEU score matter, not absolute values. Therefore, when generating training data,
hypotheses in the bottom section of the beam (according to BLEU) were assigned the target value~-1, and the rest were assigned the target value~1.

Similarly, it is only the differences in the reranker's scores that matter, and not the absolute values.
Therefore, after applying the reranker to each hypothesis in a beam, we normalise the scores to have a mean of~0.

Using the normalised BLEU scores (-1~or~1) and normalised reranker scores (with a mean of 0),
we use relative mean absolute error (RMAE) as the training objective,
as shown in Equation~(\ref{eqn:rmae}),
where $b$ is the set of hypotheses in the beam, $x$ is a hypothesis, $\hat x_{\mathrm{ranker}}$ is the normalised score predicted by the reranker, and $ x_{\mathrm{BLEU}}$ is the normalised target derived from the BLEU score ordering of the beam. 

\begin{equation}
\mathrm{RMAE}(b) =  \sum_{x \in b} | \hat x_{\mathrm{ranker}} -  \hat x_{\mathrm{BLEU}}|
\label{eqn:rmae}
\end{equation}

Several other relative loss functions have been shown to be successful in other situations \cite{zhangetal2019}. In preliminary experiments, we evaluated a number of these including log cosh error and mean square error, but they did not outperform RMAE.

\subsection{Choosing when to Manipulate}

In theory, it would be desirable to manipulate the beam at every step of hypothesis generation, but in practice, the difficulty of ranking partial hypotheses could limit its benefits. 
While manipulating the beam can avoid certain model errors, it might also introduce other errors, either from the greedy roll-out strategy or the reranker.
Reranking at every step may compound such errors.
Empirically, we found it was more effective to apply beam manipulation to some rather than all steps.


Choosing when to manipulate is thus an important decision. It is advisable to avoid manipulating the beam too early: not only it is harder to rank hypotheses with very few tokens, but it is also less likely to be beneficial. As shown in Figure~\ref{fig:fallout_location}, in the first few steps even a relatively small beam size can keep hypotheses that could lead to the reference outputs. On the other hand, it is also advisable not to manipulate too late: once hypotheses have fallen out of the beam, they cannot be put back in. As the optimal choice of when to manipulate the beam is dependent on the dataset and  the model, we treat this as a hyperparameter to be tuned on the validation set.

\section{Experiments}
\label{sec:experiments}
In this section we will present results on the E2E \cite{novikova2017e2e} and WebNLG challenges \cite{gardent-etal-2017-webnlg}. We evaluate the systems using the BLEU implementation used in the original E2E challenge.\footnote{\url{https://github.com/tuetschek/e2e-metrics}} For all the experiments reported we use the seq2seq architecture from \citet{dusek-jurcicek-2016-sequence}
for the underlying model that we are trying to manipulate.\footnote{The source code for this paper is at \url{https://github.com/jamesHargreaves12/incremental_beam_manipulation}}

It is well-known that BLEU is not a completely reliable method
for predicting human perceptions of the quality of individual NLG outputs
(for example: \citealp{callison2006bleu,novikovawe}).
However, in this case, we are comparing outputs from variants of the same system,
and thus BLEU is more likely to provide reasonable estimates of their felicity to the references,
as argued by both \citeauthor{callison2006bleu} and \citeauthor{novikovawe}.

To support the idea behind our approach, i.e.\ manipulating the beam during decoding instead of only at the end, we compare against existing rerankers applied to the beam at the end of decoding. 
These include
the TGEN reranker, proposed by \citet{dusek-jurcicek-2016-sequence},  
that has achieved state-of-the-art BLEU scores on NLG tasks, as well as   
the same reranker architecture defined in Section~\ref{sec:IR_rankers}.
Both architectures are trained 
to perform reranking of the final beam only.
When comparing between these two methods of reranking the final beam, no significant difference in performance was found.
In this section, we report results for the architecture defined in Section~\ref{sec:IR_rankers}.
For completeness, results for both rerankers are included in Appendix~\ref{sec:app_numerical_results}.

By using the same architecture both for the final-beam reranker
and for beam manipulation,
we can be more confident that any difference in results
is due to the beam manipulation strategy
(reranking during decoding, not just at the end).

In our experiments, we also consider length normalisation \cite{murray2018correcting}, which is a well-known technique that often increases the BLEU score of the resulting sequences since it addresses the bias of language models to favour short sequences, as discussed in Section~\ref{sec:relwork}. Although the values assigned to a sequence when using length normalisation are no longer probabilities, it still performs the expand, rank and prune steps at each iteration of the decoding. Hence, we can apply beam manipulation in tandem with length normalisation.
Finally,
we also considered nucleus sampling \cite{holtzman2019curious} as a baseline. However, it was found to decrease performance even when compared to vanilla beam search.

In what follows, we will not only comment on test results
when the beam size is tuned on the validation sets,
but we will also comment on test results across all beam sizes.
The reason for doing this
is that considering all beam sizes assesses whether the technique is robust to changes in beam size. In our opinion, this makes for a more convincing result then just indicating a difference in performance at a single beam size. This is especially pertinent since a well-documented issue of beam search is that larger beam sizes can lead to deteriorating performance. The results table in Appendix~\ref{sec:app_numerical_results} indicates the validation set's optimal beam size for each of the systems.

\subsection{E2E BLEU Results}
\label{sec:res_e2e}




\begin{figure}
    \centering
    \includegraphics[width=7.7cm]{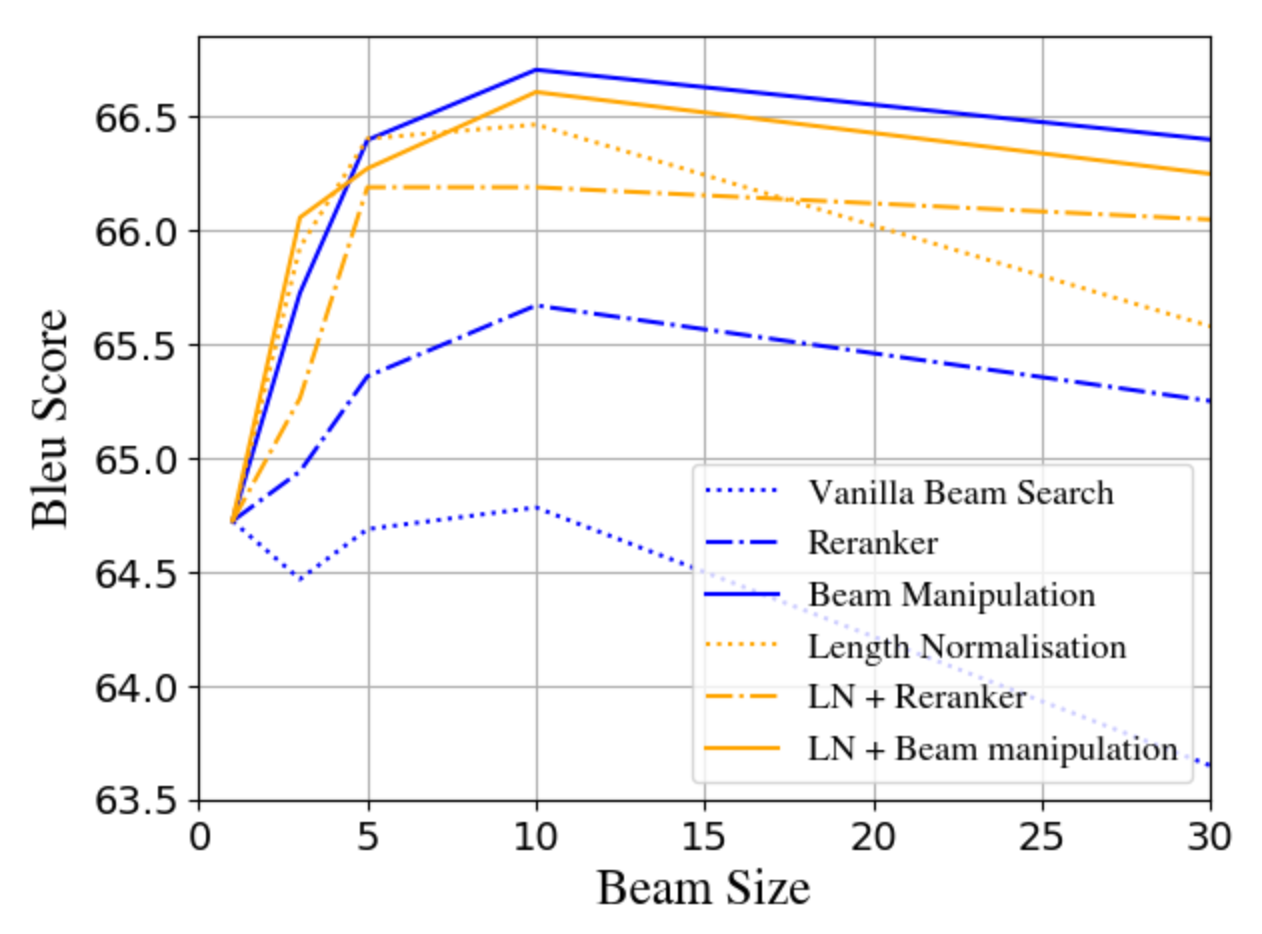}
    \caption{Results on the test set of the E2E challenge. LN~=~Length Normalisation.}
    \label{fig:e2e_test}
\end{figure}

Figure~\ref{fig:e2e_test} indicates the results on the E2E test set. 
The first thing to note is that increasing the beam size did not lead to any considerable gain in performance for the vanilla beam search strategy. For all beam sizes except 10, the performance is worse than greedy decoding, while using a beam size of 10 only increased performance by 0.06 points. Compared to vanilla beam search, reranking was an effective strategy, increasing the performance at all beam sizes. Similarly, applying incremental beam manipulation was able to outperform both methods at all beam sizes. Using the validation set to tune the beam size, the BLEU scores are 0.89 and 1.93 BLEU points higher for the reranker and incremental beam manipulation, respectively. The difference in BLEU scores between incremental beam manipulation and reranker methods was found to be significant (using a permutation test with significance level 0.01).


Length normalisation was the strongest baseline increasing the BLEU score of vanilla by 1.69 points. Adding the reranker on top of length normalisation decreases performance for all beam sizes less than 30. The strong performance of length normalisation is likely due to the fact that
the E2E test set contained longer, and more complex inputs (and hence references) than the training and validation set \cite{duvsek2020evaluating}. 
Nevertheless, applying incremental beam manipulation on top of length normalisation was able to increase the BLEU score for all beam sizes except 5.

It is worth pointing out that while incremental beam manipulation improved both vanilla beam search and length normalisation, the overall BLEU score for the combination with the latter was lower for all sizes other than size 3.
This is surprising considering that vanilla beam search performed worse than length normalisation when not combined with incremental beam manipulation. This could be due to the fact that the greedy roll-out approximation is less accurate for length normalisation than vanilla beam search since length normalisation only has an impact once some items in the beam have been completed.

\subsection{WebNLG BLEU results}
\begin{figure}
    \centering
    \includegraphics[width=7.7cm]{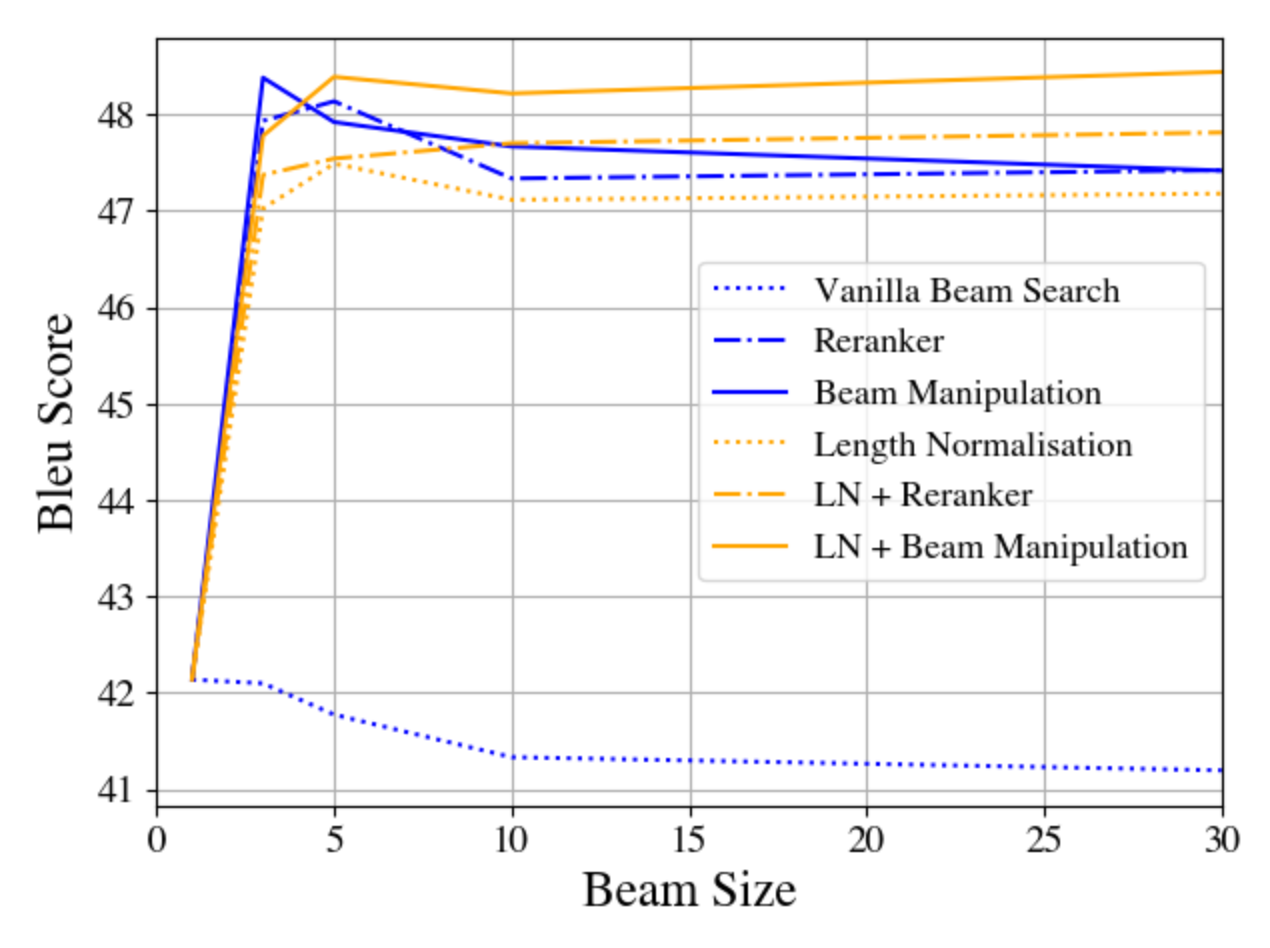}
    \caption{Results on the test set of the WebNLG challenge. LN~=~Length Normalisation.}
    \label{fig:web_test}
\end{figure}
Figure~\ref{fig:web_test} indicates the results on the WebNLG test set. 
As in the results for E2E, we can see that increasing the beam size of vanilla beam search was not an effective way to increase BLEU score. A greedy decode outperformed it at all beam sizes. Reranking the final beam was more effective, increasing the BLEU score by 5.83 points. Applying incremental beam manipulation had a very similar performance to reranking, increasing the performance at beam sizes 3 and 10 but reducing it at size 5.

The length normalisation baseline improved upon the vanilla baseline, increasing the BLEU score by 5.01 points. Reranking the final beam of the length normalised beam search was more effective on the WebNLG dataset than the E2E dataset; applying the reranker outperformed length normalisation at every beam size. Focusing on the beam sizes that performed optimally on the validation set, the BLEU score on the test set was increased by 0.43 points. Applying incremental beam manipulation on top of length normalisation received a yet higher BLEU score than reranked length normalisation for all beam sizes. Increasing the BLEU score by 1.33 points compared to the length normalisation. The improvement in BLEU scores achieved by applying incremental beam manipulation to the length normalised beam search was found to be significant when compared to length normalisation (with or without final beam reranking).
             
Unlike the E2E dataset, beam manipulation had higher performance when applied on top of length normalisation rather than vanilla beam search, outperforming it for all beam sizes except 3. The BLEU score was 0.52 points higher when taking the values at the beam sizes with the highest performances on the validation set.


 \subsection{Fallout with Beam Manipulation}
 
 In Section~\ref{sec:intro}, we explained that references often fall out a beam relatively early during decoding, and reported results on the E2E task. We repeated the same experiment for when applying incremental beam manipulation.  A beam size of 3 was used so that the results could be directly compared to those for vanilla beam search in Figure~\ref{fig:fallout_location}.
 
 \begin{figure}
    \centering
    \includegraphics[width=7.7cm]{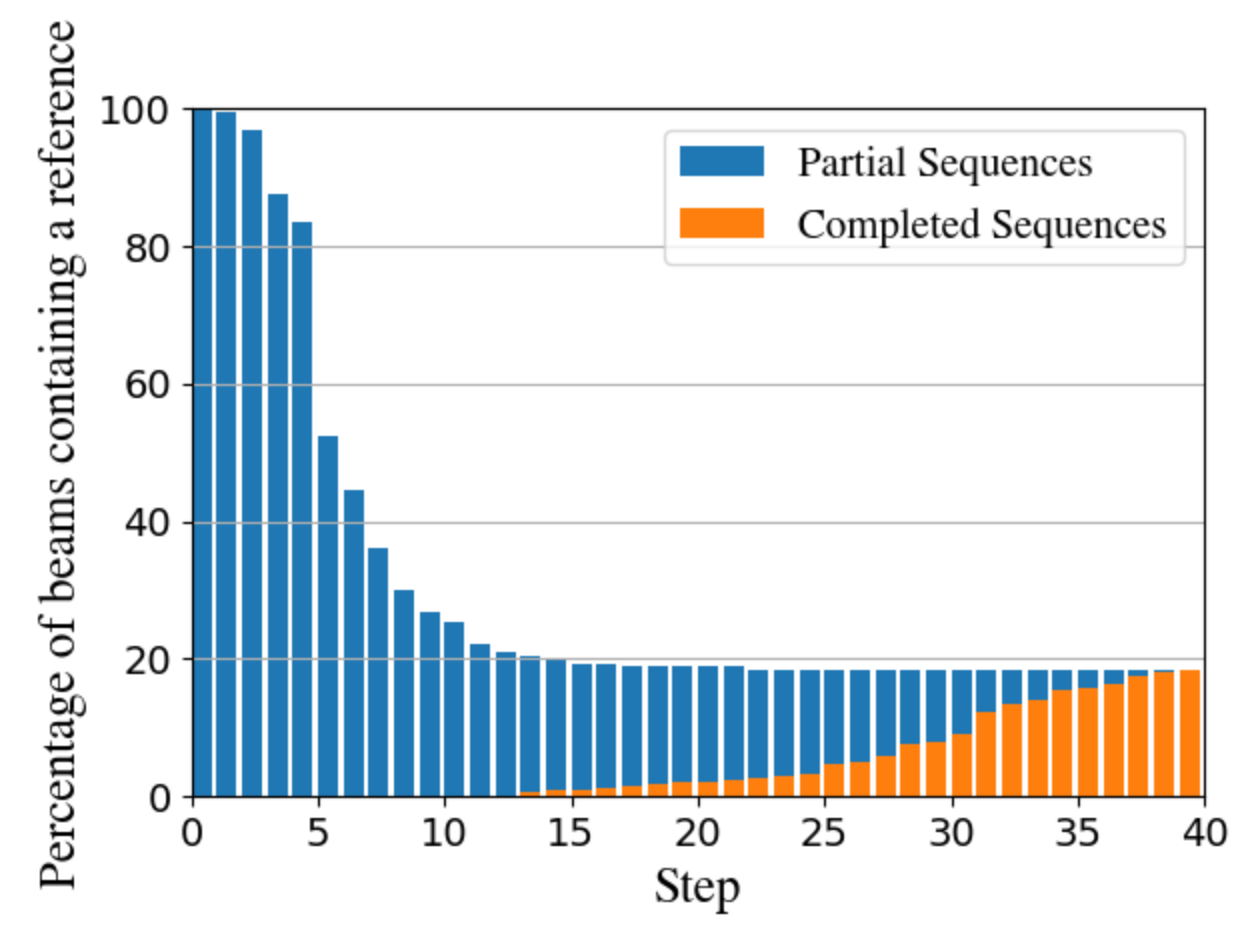}
    \caption{The percentage of beams which
    contain a reference (orange),
    or which could still lead to a reference (blue),
    using Incremental Beam Manipulation
    with beam size 3 on the E2E validation set.
    This improves on vanilla beam search, shown in Fig.~\ref{fig:fallout_location}.}
    \label{fig:fallout_bm}
\end{figure}

The results are shown in Figure~ \ref{fig:fallout_bm}.
The graph indicates that beam manipulation indeed ameliorates this issue.
The final beam contains a (correct) reference in 100/547 cases (approx 18\%), a large increase from the 60 of vanilla beam search. 
This is mainly due to reducing the number of references that fall out in steps 5 to 15, which is consistent with the fact that we are manipulating at steps 5, 10, 15 and 20. We also observed that most of the retention gain is due to earlier manipulation steps.

\subsection{Human Evaluation}
To further investigate the differences between the systems,
we conducted a human evaluation comparing the incremental beam manipulation system's output against the output of the strongest baseline on the E2E dataset -- length normalisation. 

The human evaluation was performed by the second and third authors of the paper. While the annotators had been involved in the design of the system, they had not seen textual outputs from the system prior to the annotation. The outputs were presented in a random order, without indicating which output came from which system. The systems were compared in terms of both fluency and adequacy.

For fluency, each annotator compared the system outputs for the same meaning representation input (without seeing it) and indicated their preference. Both annotators annotated 50 examples from the E2E test set.

Little difference between the outputs was found. The systems were labelled as equally fluent in 76\% of cases (incremental beam manipulation was preferred in 12\% of all cases).

To judge the adequacy of the generations, human raters were presented with the meaning representation input and the text generated by a system. They were asked to label any hallucinations and any repetitions. They were asked to ignore missing information as the human references that the system had been trained on contained frequent examples of missing information \cite{duvsek2020evaluating},
so for this dataset, missing information is better seen as content selection. Between the annotators, a combined total of 524 examples were labelled for both for hallucination and repetition.

Once again, the results were not 
conclusive in support of either system, with no statistically significant difference between them. The overall performance was very high: for 95\% of the inputs, both systems exhibited no signs of hallucination or repetition. This error analysis did, however, highlight that some errors are repeated multiple times, almost word for word. For example, all 5 cases of repetition for the incremental beam manipulation system had the following form:
``There is a pub called X located near Y. It is a pub.''

It is worth re-iterating that the system was optimised for BLEU, and not fluency or adequacy. The fact that an improvement in BLEU has not led to an improvement in a human evaluation suggests that BLEU may not be an accurate enough metric for this task, even when comparing similar systems. Therefore, BLEU may be even more limited in usefulness than \citet{callison2006bleu} and \citet{novikovawe} suggested.

\subsection{Example outputs}

\begin{figure*}[t]
\begin{subfigure}[]{\textwidth}
\centering
\begin{tabular}{|p{0.95\textwidth}|}
\hline
\textbf{Input:} name = The Cricketers $\vert$ eat type = restaurant $\vert$ food = English $\vert$ price range = cheap $\vert$\linebreak rating = average $\vert$ area = city centre $\vert$ family friendly = yes $\vert$ near = Café Rouge \\
\hline
\textbf{BM:}     The Cricketers serves cheap English food in the city centre near Café Rouge. It has an average customer rating and is family-friendly. \\
\textbf{LN:}     The Cricketers is a cheap, family-friendly, English restaurant with an average customer rating. It is located in the city centre near Café Rouge and has an average customer rating. \\
\textbf{RR:}     The Cricketers is a cheap, family-friendly restaurant located in city centre near Café Rouge. \\ 
\hline
\end{tabular}
\end{subfigure}

\vspace{\abovecaptionskip}

\begin{subfigure}[]{\textwidth}
\centering
\begin{tabular}{|p{0.95\textwidth}|}
\hline
\textbf{Input:} name = The Phoenix $\vert$ eat type = pub $\vert$ food = French $\vert$ price range = £20-25 $\vert$ rating = 3 out of 5 $\vert$ area = riverside $\vert$ family friendly = no $\vert$ near = Café Sicilia\\
\hline
\textbf{BM:}     The Phoenix is a French restaurant in riverside near Café Sicilia. It has a moderate price range and a customer rating of 3 out of 5. It is not kid friendly. \\
\textbf{LN:}     The Phoenix is a restaurant providing French food in the £20-25 price range. It is located in the riverside. It is near Café Sicilia. Its customer rating is 5 out of 5. \\
\textbf{RR:}     The Phoenix is a restaurant providing French food in the £20-25 price range. It is located in the riverside. It is near Café Sicilia. Its customer rating is high. \\ 
\hline
\end{tabular}
\end{subfigure}



\caption{Example outputs for different systems. BM = Incremental beam manipulation system, LN = Vanilla beam search with length normalisation, RR = Vanilla beam search with reranking applied to the final beam.}
\label{example-outputs}
\end{figure*}

We now present a couple of examples where manipulating the beam during decoding led to an improvement in the quality of the output. These were selected from the set of examples for which the output of the beam manipulator was preferred by the human annotators in terms of adequacy. The examples are given in Figure~\ref{example-outputs}.

In the first example, we can see that length normalisation leads to a repetition of the fact that The Cricketers had an `average customer rating', showing the downsides of a technique that just favours longer outputs. Neither the output from the beam manipulator nor the reranked approach contain repetitions,  although we can see that more of the input information is realised in the case of the beam manipulator.

The second example contains a hallucination for both the length normalised and reranked systems -- the input clearly states that the customer rating was `3 out of 5'. In contrast, whereas these systems claim that it was `5 out of 5' and `high' respectively. The beam manipulation system avoided this issue.

\section{Conclusions}
Rerankers are commonly used to increase the performance of NLG systems decoded by beam search, by modifying which hypothesis from the final beam is chosen. This means that rerankers are dependent on good hypotheses reaching the final beam. However, this is often not the case; on the validation set of E2E challenge, only 11\% of references were present in the final beam when the seq2seq model from \citep{dusek-jurcicek-2016-sequence} was decoded with a beam size of 3.

To address this limitation, we proposed incremental beam manipulation,
which modifies the ranking of partial hypotheses within the beam at intermediate steps of the decoding, and hence chooses which are pruned. We evaluated this method on both the E2E and WebNLG challenges. The results showed that applying beam manipulation, instead of a reranker, was able to increase the BLEU score by 1.04 on the E2E challenge. We further showed that incremental beam manipulation was able to increase performance when applied on top of length normalisation.

The optimal reranker for incremental beam manipulation may differ at each step of generation (for example, token 5 vs.\ token 20). In future work, we intend to refine our method further by conditioning the reranker on how far through the beam search we are.

\bibliography{anthology,references}
\bibliographystyle{acl_natbib}

\appendix

\section{Fallout experiment with larger beam size}
\label{sec:app_fallout}
\begin{figure}
    \centering
    \includegraphics[width=7.7cm]{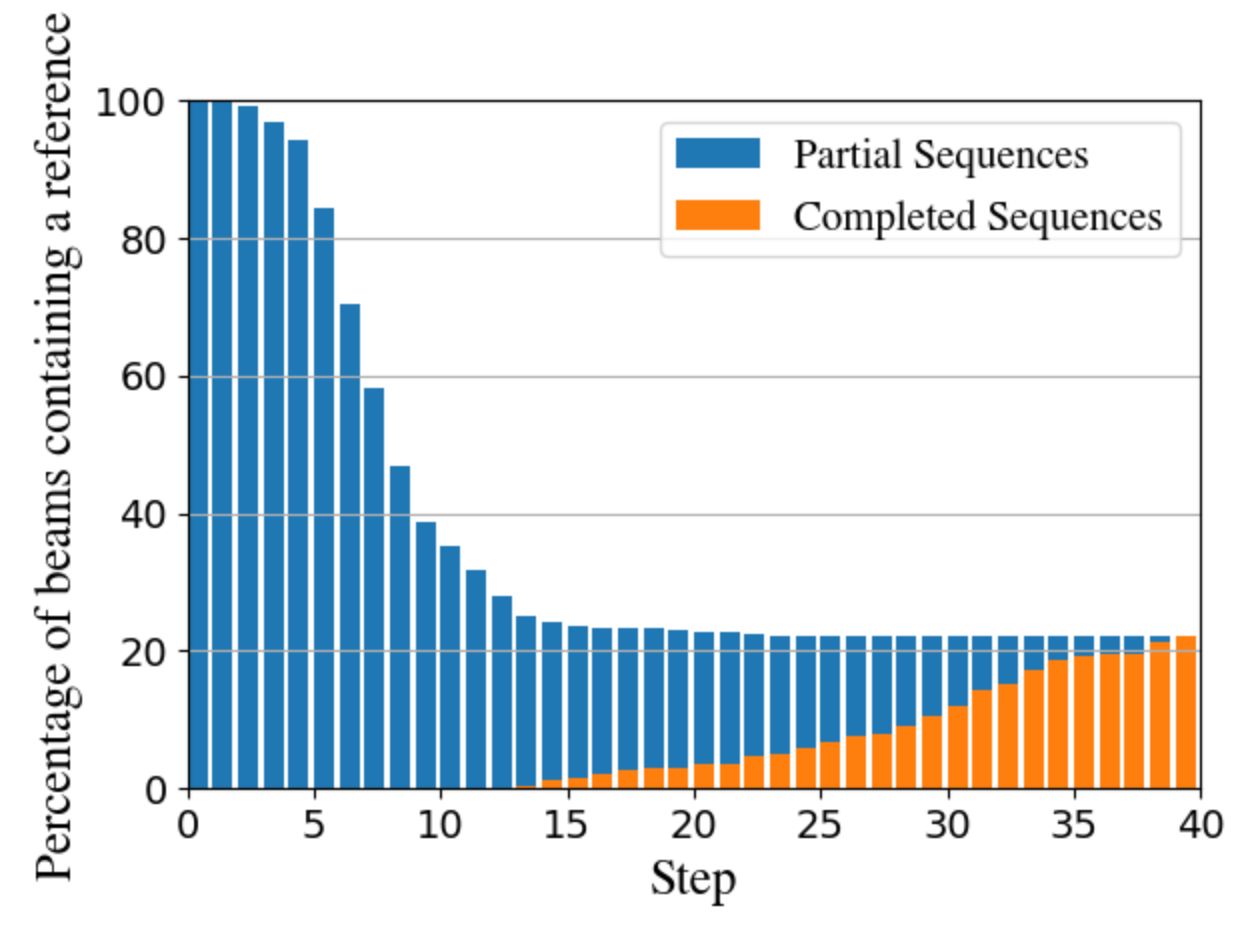}
    \caption{The percentage of beams that contain a reference sentence after each step of beam search. A beam size of 10 was used to decode the model proposed by \citet{dusek-jurcicek-2016-sequence}. Results are for the E2E validation dataset. The orange bars indicate the number of completed references within the beam.}
    \label{fig:fallout_location_10}
\end{figure}

Figure~\ref{fig:fallout_location} indicates the step at which the reference sentences drop out of the beam (for a beam size of 3). Figure~\ref{fig:fallout_location_10} indicates the same results for a larger beam size of 10. 

The figure indicates that the number of references that were contained in the final beam was higher for a beam size 10. For the early iterations of the decoding the number of references that fell out of the beam was far lower for a beam size of 10. A larger beam size meant that the beam contained more hypotheses and so has more chances to match against a reference.

However the shape of the graphs is very similar. The majority of references that fell out did so relatively early in the process. 54\% of references fell out by step 7, increasing to 79\% by step 9. At step 21 the final last reference fell out of the beam despite the fact that the beam contained partially references up to step 40. 
 
\section{Pointwise vs Pairwise rerankers}
\label{sec:app_point_pair}
This paper required a method of ranking completed hypotheses from worst to best. During preliminary experiments we implemented rerankers based on the Pairwise and Pointwise strategies from the Information Retrieval field. See Section~\ref{sec:IR_rankers} for more details.
 
\begin{figure}
    \centering
    \includegraphics[width=7.7cm]{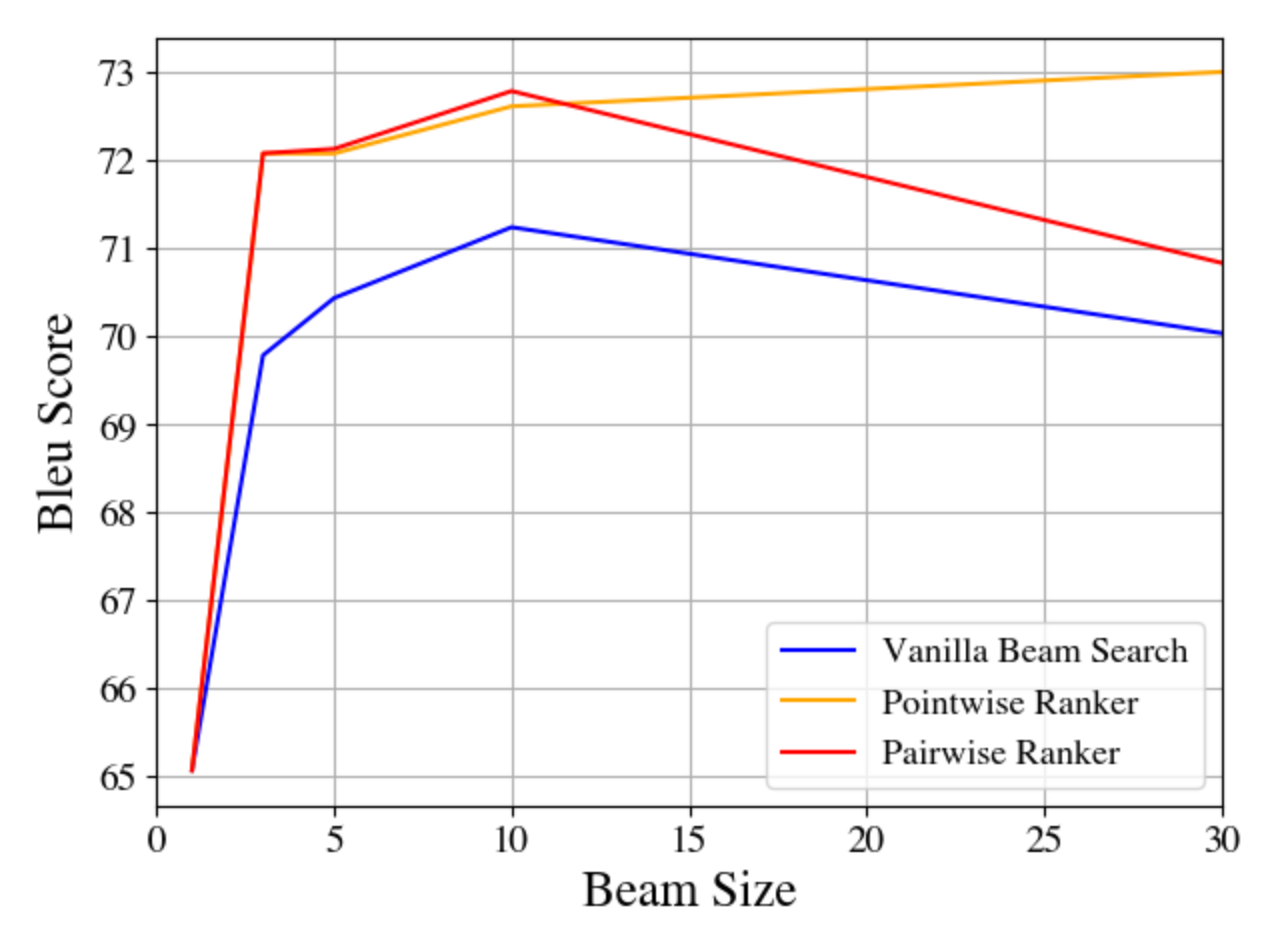}
    \caption{Comparison between the performance of the pointwise and pairwise rankers when used as rerankers on the E2E validation set.}
    \label{fig:point_vs_pair}
\end{figure}

To evaluate the performance of the different rankers we applied each of the rankers as a reranker of the final beam of a vanilla beam search over the E2E validation set. The BLEU scores for each of the rerankers were calculated for each beam size. The results are shown in Figure \ref{fig:point_vs_pair}.

We can see that there was very little difference in performance for the two methods of reranking for the beam sizes up to 10. However, for beam size 30 the pointwise reranker significantly outperforms the pairwise reranker. The larger the beam size the greater the number of hypotheses that the reranker can pick as top and hence the greater the impact of the reranker. 

The pointwise reranker requires O(k) runs of the reranker to produce a total ordering. On the other hand the Copeland method to produce a total ordering from the pairwise comparisons requires $O(k^2)$ number of pairwise Comparisons.

These factors lead us to choose the Pointwise ranker over the pairwise ranker for the experiments in the results section.

\begin{table*}[]
\centering
\begin{tabular}{l|lllllll}
\textbf{Beam size} & \textbf{Vanilla} & \textbf{Rerank} & \textbf{TGEN} & \textbf{LN}    & \textbf{LN+Rerank} & \textbf{BM}     & \textbf{LN+BM} \\ \hline
1                  & 64.72            & 64.72           & 64.72         & 64.72          & 64.72              & 64.72           & 64.72          \\
3                  & 64.47            & 64.94           & 65.33         & 65.93          & 65.26              & 65.73           & \textbf{66.06} \\
5                  & 64.69            & 65.36           & 65.47*        & \textbf{66.40} & 66.19*             & \textbf{66.40}  & 66.27          \\
10                 & 64.78*           & 65.67*          & 65.58         & 66.47*         & 66.19              & \textbf{66.71}* & 66.61          \\ 
30                 & 63.65            & 65.25           & 65.44         & 65.58          & 66.05              & \textbf{66.40}  & 66.25*         
\end{tabular}
\caption{BLEU scores for each of the different systems on the E2E testset. *indicates the beam size which scored highest on the respective validation sets.\textbf{bold} indicates the highest scoring system for each beam size.}
\label{tab:e2e_res}
\end{table*}

\begin{table*}[]
\centering
\begin{tabular}{l|lllllll}
\textbf{Beam size} & \textbf{Vanilla} & \textbf{Rerank} & \textbf{TGEN} & \textbf{LN} & \textbf{LN+Rerank} & \textbf{BM}    & \textbf{LN+BM} \\ \hline
1                  & 42.14            & 42.14           & 42.14         & 42.14       & 42.14              & 42.14          & 42.14          \\
3                  & 42.10*           & 47.93*          & 47.28*        & 47.02       & 47.37              & \textbf{48.38} & 47.78          \\
5                  & 41.77            & 48.13           & 47.41         & 47.49       & 47.54*             & 47.92*          & \textbf{48.39} \\
10                 & 41.33            & 47.33           & 46.50         & 47.11*      & 47.70              & 47.66         & \textbf{48.21} \\
30                 & 41.20            & 47.42           & 46.61         & 47.18       & 47.81              &   47.41        & \textbf{48.44}*
\end{tabular}
\caption{BLEU scores for each of the different systems on the WebNLG testset. *indicates the beam size which scored highest on the respective validation sets. \textbf{bold} indicates the highest scoring system for each beam size.}
\label{tab:web_res}
\end{table*}

\section{Numerical results}
\label{sec:app_numerical_results}
This section will present the numerical results for the E2E and WebNLG datasets so that they can be more readily compared in future works. The results are given in Table \ref{tab:e2e_res} and Table \ref{tab:web_res}.

\section{Hyperparameters}
Throughout this paper a number of hyperparameters were introduced. The values used for each of the models in this paper are summarised bellow. Note that the search for these values was far from exhaustive so there is a good chance that the results of this paper could be improved upon through a better optimisation procedure. 

In Section~\ref{sec:training_ranker}, the beam is split into two sections, bottom and rest. For all beam manipulation models the bottom of the beam was set to the bottom (i.e.~lowest scoring) quarter of the beam. We also say a large beam is used to generate the data for training the beam. For all experiments in this paper we use a beam size of 50.

A key hyperparameter for performance of the incremental beam manipulation was the steps of the beam search at which the beam was manipulated. This hyperparameter varied for the 4 separate beam manipulation models. The values are summarised as follows:
\begin{itemize}
    \item E2E, Incremental Beam Manipulation on top of vanilla beam search: 5, 10, 15, 20 and final.
    \item E2E, Incremental Beam Manipulation on top of length normalised beam search: 5, 7 and 10.
    \item WebNLG, Incremental Beam Manipulation on top of vanilla beam search: 4, 12 and final.
    \item WebNLG, Incremental Beam Manipulation on top of length normalised beam search: 5 and 12.
\end{itemize}

It is worth noting that manipulating the final step is the same as reranking the beam according to the ranker used in beam manipulation (i.e. no rollouts are performed).



\end{document}